\documentclass{article}

\usepackage{microtype}
\usepackage{graphicx}
\usepackage{subcaption}
\usepackage{booktabs} 

\usepackage{hyperref}

\usepackage[preprint]{icml2026}

\makeatletter
\renewcommand{\printAffiliationsAndNotice}[1]{%
  \global\icml@noticeprintedtrue%
  {\let\thefootnote\relax\footnotetext{\hspace*{-\footnotesep}#1\ \\
      \Notice@String}}%
}
\makeatother

\usepackage{amsmath}
\usepackage{amssymb}
\usepackage{mathtools}
\usepackage{amsthm}
\usepackage{algorithm}
\usepackage{algorithmic}
\usepackage{multirow}

\usepackage[capitalize,noabbrev]{cleveref}

\icmltitlerunning{SeeMe: Mitigating Hallucinations in LVLMs through Visual Token Engineering}

\begin{document}

\twocolumn[
  \icmltitle{SeeMe: Mitigating Hallucinations in Large Vision-Language Models \\
    through Effective Visual Token Engineering}


  \begin{center}
    {\bf Kai Tang}\textsuperscript{1,4,*}\quad
    {\bf Jinhao You}\textsuperscript{2,*}\quad
    {\bf Bohua Zhang}\textsuperscript{1,3}\quad
    {\bf Yichen Guo}\textsuperscript{1,4}\quad
    {\bf Yiding Sun}\textsuperscript{1}\quad
    {\bf Dongxu Zhang}\textsuperscript{5}\quad
    {\bf Chenxi Li}\textsuperscript{6} \\
    {\bf Xiande Huang}\textsuperscript{7,\ensuremath{\dagger}}\quad
    {\bf Shanghang Zhang}\textsuperscript{1,\ensuremath{\dagger}} \\
    \small
    \textsuperscript{1}\,State Key Laboratory of Multimedia Information Processing, \\
    School of Computer Science, Peking University \\
    \textsuperscript{2}\,University of Pennsylvania \\
    \textsuperscript{3}\,University of Electronic Science and Technology of China \\
    \textsuperscript{4}\,Nanyang Technological University \\
    \textsuperscript{5}\,Tsinghua University \\
    \textsuperscript{6}\,The Chinese University of Hong Kong, Shenzhen \\
    \textsuperscript{7}\,De Artificial Intelligence Lab \\
    \footnotesize \textsuperscript{*}Equal contribution.
    \textsuperscript{\ensuremath{\dagger}}Corresponding authors.
  \end{center}

  \icmlkeywords{Vision-Language Models, Hallucination, Visual Token Engineering, Machine Learning}
  \hypersetup{pdfauthor={Kai Tang, Jinhao You, Bohua Zhang, Yichen Guo, Yiding Sun, Dongxu Zhang, Chenxi Li, Xiande Huang, Shanghang Zhang}}

  \vskip 0.2in
]

\printAffiliationsAndNotice{Contact: Kai Tang \textless{}kaitang030113@gmail.com\textgreater{};
Correspondence: Shanghang Zhang \textless{}shanghang@pku.edu.cn\textgreater{}.}

\begin{abstract}

Large Vision-Language Models (LVLMs) have achieved remarkable progress in visual understanding tasks such as image captioning and visual question answering. However, they remain susceptible to hallucinations, generating content that is inconsistent with the actual visual input. Existing methods primarily intervene at the decoding stage, while overlooking a critical source of hallucinations: irrelevant or noisy visual tokens that mislead the decoding process. To address this issue, we propose \textbf{SeeMe}, a training-free framework that introduces the concept of feature engineering from traditional machine learning into LVLMs. SeeMe restructures visual tokens through a three-stage token engineering process to suppress hallucination sources while preserving informative visual evidence. Experiments on MME, POPE, and AMBER benchmarks across four LVLMs demonstrate that SeeMe consistently reduces hallucinations and improves output consistency, providing a novel perspective for mitigating hallucinations in LVLMs.

\end{abstract}

\section{Introduction}

\begin{figure}[t]
\centering
\includegraphics[width=1\linewidth]{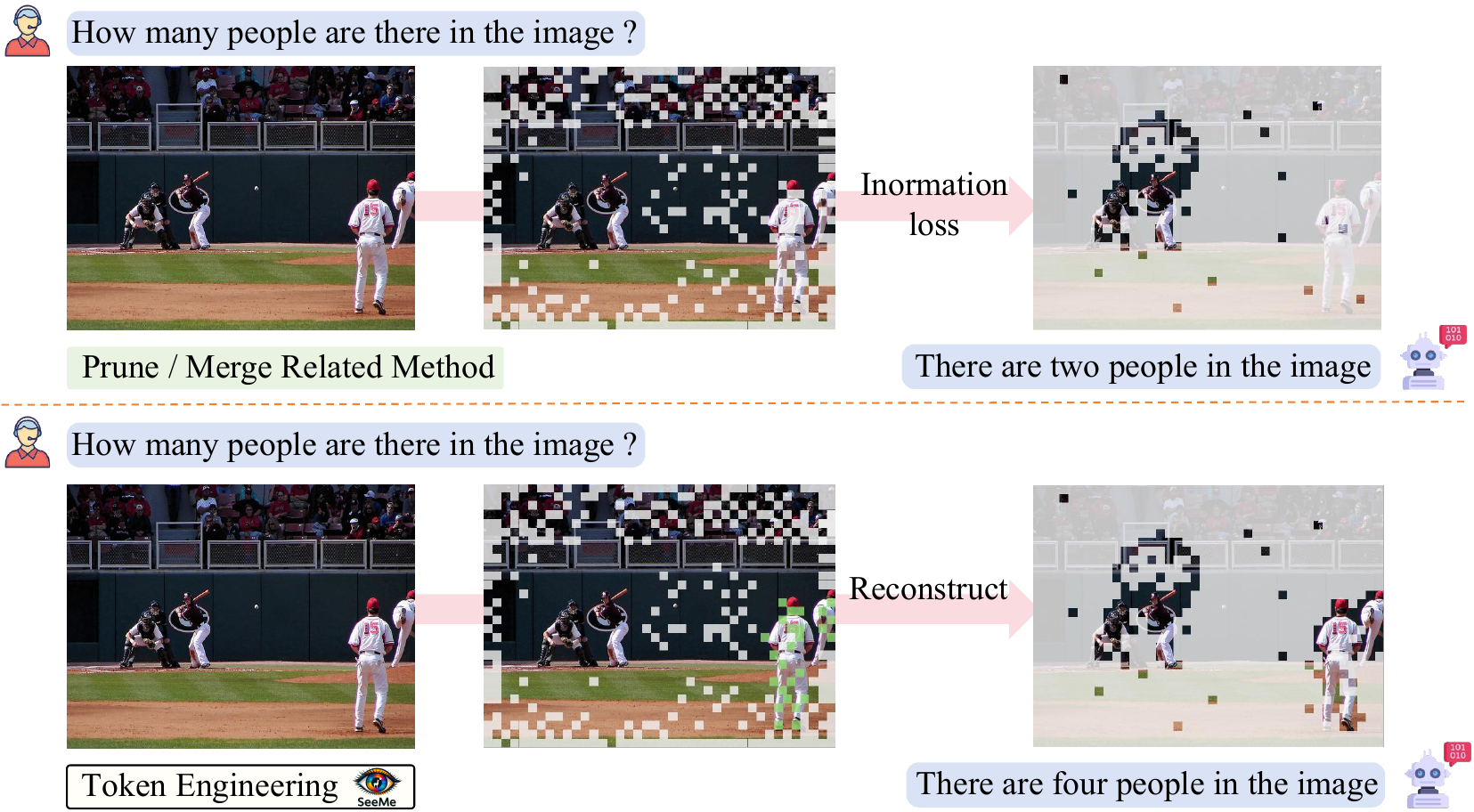}
\caption{Comparison of visual token handling strategies. Token engineering reconstructs informative representations, leading to more accurate answers.}
\label{fig_intro}
\vspace{-3mm}
\end{figure}

Large Vision-Language Models (LVLMs) have achieved remarkable success in open-ended visual understanding tasks, including image captioning, visual question answering, and multimodal dialogue \citep{liu2023llava, dai2023instructblip, zhu2023minigpt4, bai2023qwen, ye2024mplug}. Despite these advances, LVLMs remain prone to hallucination—generating content that is inconsistent with the actual visual input \citep{liu2024survey}. This phenomenon severely undermines their reliability in practical applications, particularly in safety-critical or factual scenarios \citep{hartsock2024vision,zhou2024vision,ma2024survey}.

Recent studies have attributed hallucinations in LVLMs to several factors, including over-reliance on statistical biases in training data \citep{zhou2023analyzing, chen2024multi}, the dominance of language priors over weak visual grounding \citep{han2022visual, guan2024hallusionbench}, and the dilution of cross-modal attention in deeper layers \citep{an2025mitigating}. To mitigate these issues, recent work has explored training-free strategies that intervene during inference. These methods typically operate on the language decoder's internal dynamics: DoLa \citep{chuang2023dola} compares early and late layer outputs, VCD \citep{leng2024mitigating} contrasts original and distorted visual inputs, DAMO \citep{wang2025damo} accumulates past activations to stabilize hidden states, and DCLA \citep{tang2026mitigatinghallucinationsinterlayerconsistency} enforces inter-layer consistency to reduce semantic drift.

Although these methods have shown effectiveness, they primarily operate on the language decoder's internal dynamics, overlooking a key factor: visual tokens themselves are often a primary source of hallucinations. In many cases, hallucinations arise from irrelevant or noisy visual tokens passed from the visual encoder to the language decoder, leading the model to generate outputs inconsistent with the actual visual content \citep{che2025eazy, woo2024don, an2025mitigating}. Recent studies have addressed this issue by directly manipulating visual tokens: EAZY \citep{che2025eazy} identifies hallucinatory tokens through attention analysis and zeros them out, while SPIN \citep{sarkar2025mitigating} suppresses attention heads that exhibit low attention to image tokens.

However, these suppression-based methods inherently face a trade-off between hallucination mitigation and semantic preservation: aggressive suppression effectively removes noise but risks discarding useful visual evidence. To address this challenge, we propose \textbf{SeeMe}, a training-free framework that \textit{restructures} visual tokens through a three-stage token engineering process, rather than merely suppressing them. The first stage, \textbf{Se}l\textbf{e}ction, prunes irrelevant visual tokens using cross-modal attention. In the \textbf{Me}rging stage, similarity-driven fusion is used to combine visual tokens from the original input with similar semantics, forming a high-quality token pool. The final \textbf{Se}l\textbf{e}ction stage refines the merged tokens by selecting those that best align with the language context, effectively complementing the initial pruning step. Through this progressive restructuring, SeeMe improves visual grounding while reducing hallucination and redundancy, without requiring any retraining or architectural changes.

Experiments on MME, POPE, and AMBER benchmarks demonstrate that SeeMe consistently reduces hallucinations across four mainstream LVLMs: LLaVA-1.5 \citep{liu2023llava}, LLaVA-NEXT \citep{liu2024llavanext}, INF-MLLM \citep{zhou2023infmllm}, and mPLUG-Owl2 \citep{ye2024mplug}, without requiring additional training. The results highlight SeeMe's strong generalizability and effectiveness in hallucination mitigation.

\textbf{Contributions.} Our main contributions are as follows:
\begin{itemize}
    \item We introduce the concept of \textit{feature engineering} from traditional machine learning into LVLM hallucination mitigation for the first time, proposing to actively restructure visual tokens rather than merely suppressing them.
    \item We present SeeMe, a training-free, three-stage framework that first filters irrelevant tokens, then enriches the token pool through semantic fusion, and finally refines the representation by selecting high-quality candidates aligned with textual context.
    \item Extensive experiments on MME, POPE, and AMBER benchmarks across four LVLMs (LLaVA-1.5, LLaVA-NEXT, INF-MLLM, mPLUG-Owl2) demonstrate that SeeMe consistently improves hallucination mitigation, highlighting its effectiveness and generalizability.
\end{itemize}

\section{Related Work}
\subsection{Large Vision-Language Models}
Large Vision-Language Models (LVLMs) have evolved from BERT-style multimodal encoders \citep{lu2019vilbert,devlin2019bert,liu2019roberta} to decoder-based architectures powered by Large Language Models (LLMs) \citep{touvron2023llama, chiang2023vicuna}. In this paradigm, visual inputs are encoded into tokens and passed to a pretrained LLM, enabling unified decoding across modalities. End-to-end training methods \citep{jia2021scaling,radford2021learning} have improved cross-modal alignment, while instruction-tuned models such as LLaVA \citep{liu2023llava} and InstructBLIP \citep{dai2023instructblip} have further enhanced performance on open-ended vision-language tasks. Recent work has also explored efficient and explicit reasoning for multimodal systems, including chain-of-thought compression, adaptive coarse-to-fine refinement, late-stage fragility analysis, and 3D geometric reasoning benchmarks \citep{zhang2026chain,zhang2026not,zhang2025not,zhang2026pointcot}.

\subsection{Hallucination in LVLMs}
Hallucination—generating content that is inconsistent with the source input—has long been a concern in natural language generation \citep{yao2023llm,zhu2025ibd,xu2024hallucination}. With the rise of LVLMs, this problem becomes more prominent, as the model must align high-dimensional visual inputs with text generation in a multimodal setting \citep{liu2024survey,ye2024mplug,liu2023llava}. Hallucinations in LVLMs typically manifest as descriptions of non-existent objects, incorrect attributes, or globally plausible but visually unfaithful outputs \citep{liu2023mitigating,li2023evaluating}.

Prior approaches to mitigating hallucinations typically rely on supervised fine-tuning or data augmentation \citep{wang2024mitigating,xiao2025detecting}, or train correction modules to detect and revise hallucinated content \citep{liu2023mitigating, gunjal2024detecting}. However, these solutions require additional annotation and computation, limiting their scalability. Recent works have explored training-free mitigation strategies at inference time, such as DCLA \citep{tang2026mitigatinghallucinationsinterlayerconsistency}, which enforces inter-layer consistency, DAMO \citep{wang2025damo}, which accumulates internal activations to stabilize hidden states, and FADE \citep{guo2026fademitigatinghallucinationsreducing}, which reduces language-prior dominance. Yet these methods operate entirely on the language decoder's internal dynamics.

\begin{figure*}[t]
\centering
\includegraphics[width=0.9\textwidth]{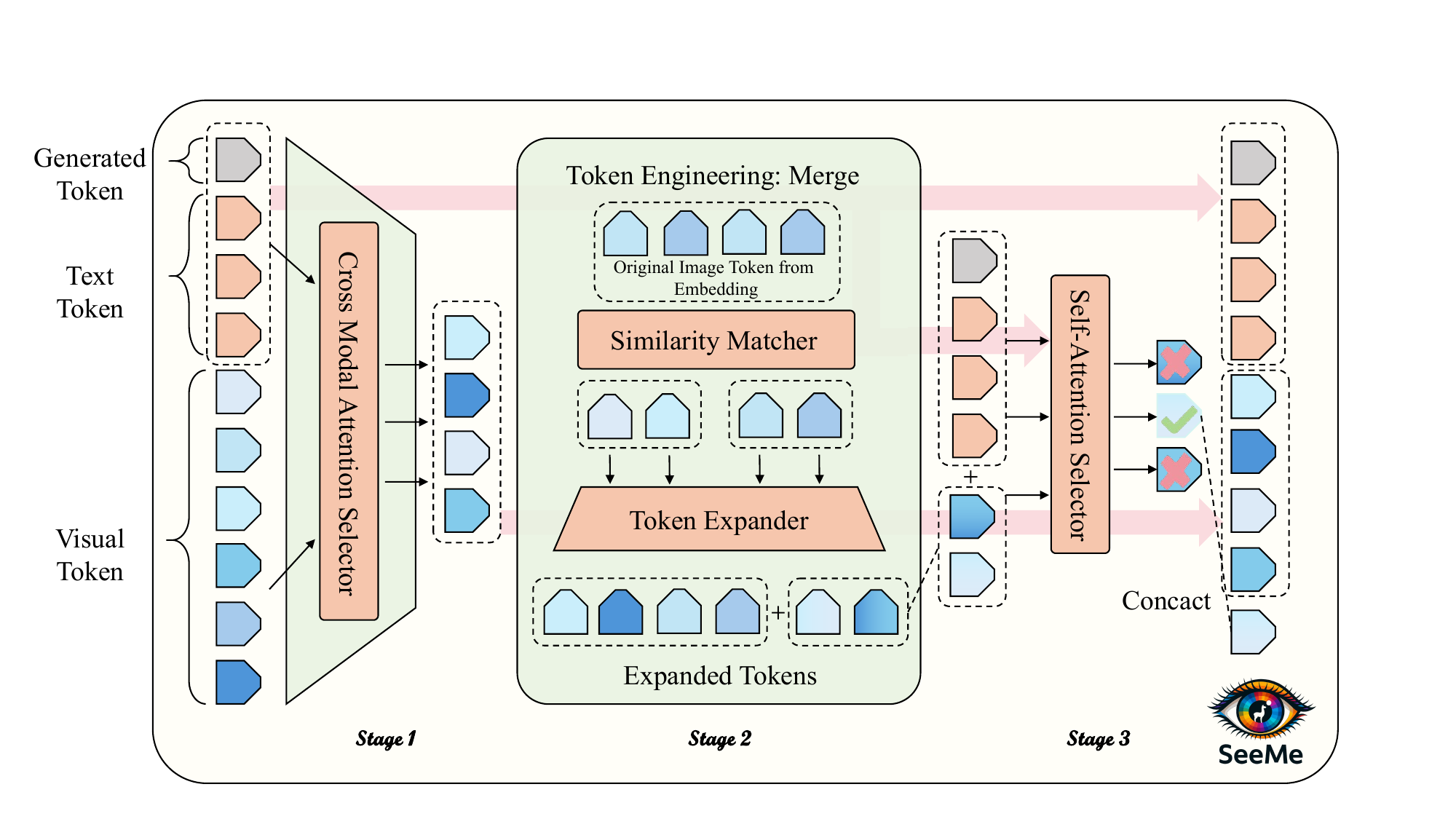}
\caption{Architecture of the proposed SeeMe framework, which restructures visual tokens through three stages: cross-modal pruning, semantic merging, and self-attention-based refinement.}
\label{fig_method}
\vspace{-3mm}
\end{figure*}

\subsection{Visual Token Manipulation}
In typical LVLM architectures, an image is processed by a vision encoder into a sequence of visual tokens, which are then passed to the language decoder \citep{radford2021learning, dai2023instructblip}. A $336 \times 336$ image typically yields 576 visual tokens, and higher-resolution images can produce several thousand \citep{bolya2022token,shang2024llava,kim2024token}. Prior studies have found that many of these tokens receive minimal attention from text queries and contribute little to the final output \citep{zhang2024sparsevlm, guo2025star, chen2024image}.

To address this inefficiency, token reduction methods have been proposed. ToMe \citep{bolya2022token} merges similar tokens to reduce computation, while STAR \citep{guo2025star} prunes visual tokens based on cross-modal attention for efficient inference. More recently, methods have directly targeted hallucination through token manipulation: EAZY \citep{che2025eazy} identifies hallucinatory tokens via attention analysis and zeros them out, while SPIN \citep{sarkar2025mitigating} suppresses attention heads with low image attention. However, these suppression-based approaches face a fundamental trade-off between hallucination mitigation and semantic preservation—aggressive suppression risks discarding useful visual evidence.

\section{Motivation}
\begin{figure}[h]
\centering
\includegraphics[width=0.9\linewidth]{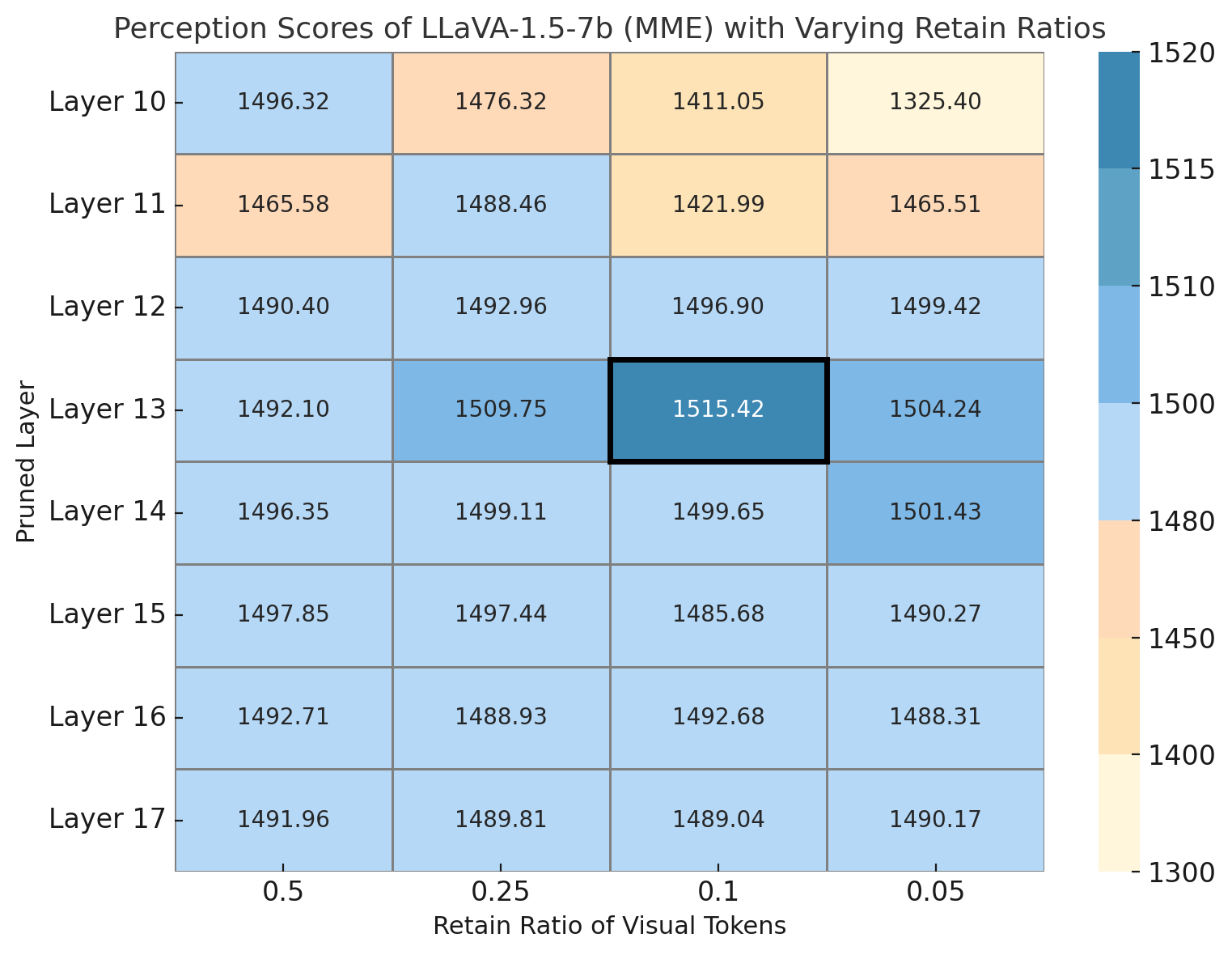}
\caption{Perception performance of LLaVA-1.5-7B on the MME dataset with varying retain ratios of visual tokens at different intermediate layers.}
\label{fig_motivation}
\vspace{-5mm}
\end{figure}

To investigate the relationship between visual token manipulation and hallucination, we conducted a preliminary experiment in which we pruned visual tokens based on cross-modal attention at different decoder layers. The results in \cref{fig_motivation} show that pruning visual tokens in middle layers significantly reduces hallucinations, but excessive pruning causes information loss and degrades performance. To avoid loss of visual information, we design a visual token engineering process to construct an informative visual token pool, from which task-relevant tokens are further selected to supplement the visual token sets.

Based on this observation, we design \textbf{SeeMe} to address hallucination at its root by restructuring visual tokens rather than merely suppressing them.

\section{Method}
Inspired by the classical philosophy of feature engineering \citep{zhang2023openfe}, SeeMe treats visual tokens as editable features and applies a three-stage editing process within the frozen decoder of an LVLM. As shown in \cref{fig_method}, SeeMe consists of the following steps: (1) an initial \textbf{Selection} stage prunes semantically irrelevant visual tokens using cross-modal attention; (2) a \textbf{Merging} stage fuses locally similar tokens through similarity-weighted aggregation to recover fine-grained evidence; and (3) a final attention-guided \textbf{Selection} retains only the most linguistically aligned fused tokens. This progressive restructuring not only reduces redundancy, but also generates high quality, semantically faithful visual representations—ultimately lowering the risk of hallucination without additional training or architectural changes.

\subsection{Stage 1: Cross-Modal Attention Selector}
To reduce unnecessary computation and suppress hallucination sources early in the pipeline, we introduce a cross-modal attention-based filtering mechanism to eliminate semantically irrelevant visual tokens before further processing. We adopt a pruning strategy inspired by STAR \citep{guo2025star}. Specifically, we perform token filtering after a decoder layer where visual-textual attention has matured. This allows the model to first attend to relevant visual details and establish cross-modal grounding. By pruning at this point, we remove semantically irrelevant tokens while preserving those essential for grounding, thereby reducing the risk of hallucination in the language generation stage.

Let $H_v \in \mathbb{R}^{L_v \times d}$ denote the visual token embeddings, and let $H_q \in \mathbb{R}^{L_q \times d}$ and $H_{\text{resp}} \in \mathbb{R}^{L_o \times d}$ represent the embeddings of the input query and the generated response, respectively. We concatenate the textual components to form:
\[
\hat{H}_q = [H_q; H_{\text{resp}}] \in \mathbb{R}^{(L_q + L_o) \times d}.
\]

At decoder layer $K$,  we extract the cross‐modal attention weights produced by the model at layer $K-1$:
\[
C_{K-1} = \text{Softmax} \left( \frac{\hat{H}_q H_v^\top}{\sqrt{d}} \right) \in \mathbb{R}^{(L_q + L_o) \times L_v}.
\]

We then compute an importance score $r_i$ for each visual token $i$ by averaging its attention weights over all textual tokens:
\[
r_i = \frac{1}{L_q + L_o} \sum_{j=1}^{L_q + L_o} C_{K-1}[j, i], \quad \text{for } i = 1, \ldots, L_v.
\]

This results in an importance vector $\vec{r} = [r_1, r_2, \ldots, r_{L_v}] \in \mathbb{R}^{L_v}$. We select the top-$k$ visual tokens according to their scores, where $k = \lfloor P \cdot L_v \rfloor$ and $P \in (0, 1)$ is a predefined retention ratio:
\[
\text{RetainedIndices} = \text{TopK}(\vec{r}, k).
\]

We apply this hard pruning by directly removing the unselected visual tokens from the sequence, while maintaining the relative order of the remaining ones. This design improves efficiency and helps reduce hallucination by filtering out visually irrelevant context early in the decoding process.

\begin{algorithm}[t]
\caption{\textbf{SeeMe}}
\label{alg:seeme}
\textbf{Input:} Visual tokens \( \mathbf{Z}_v \), system prompt \( \mathbf{X}_s \), textual query \( \mathbf{X}_q \) \\
\textbf{Output:} Restructured token sequence \( \tilde{\mathbf{Z}} \)

\textbf{Stage 1: Cross-modal Selection} \\
Obtain attention map \( \mathbf{A} \) at decoder layer \( K-1 \) \\
Compute cross-modal attention scores from \( [\mathbf{X}_q; \mathbf{X}_{\text{resp}}] \) to \( \mathbf{Z}_v \) \\
Select top-\( k_1 = r \cdot |\mathbf{Z}_v| \) visual tokens as \( \mathbf{Z}_{v}^{\text{sel}} \)

\textbf{Stage 2: Similarity-guided Token Fusion} \\
Normalize original \( \mathbf{Z}_v \) from embedding layer, compute cosine similarity \\
For each token, retrieve \( k \)-nearest neighbors \\
Merge token with neighbors using similarity-weighted fusion \( \rightarrow \mathbf{Z}_v^{\text{fused}} \) \\
Concatenate: \( \mathbf{Z}_v^{\text{enh}} = [\mathbf{Z}_v; \mathbf{Z}_v^{\text{fused}}] \)

\textbf{Stage 3: Final Attention-based Selection} \\
Compute cross-modal attention from text to \( \mathbf{Z}_v^{\text{enh}} \) \\
Select top-\( n \) tokens \( \mathbf{Z}_v^{\text{final}} \) by attention score

\textbf{Output:} Concatenate final sequence: \\
\( \tilde{\mathbf{Z}} = [\mathbf{X}_s; \mathbf{Z}_v^{\text{final}}; \mathbf{Z}_v^{\text{sel}}; \mathbf{X}_q] \)
\end{algorithm}

\subsection{Stage 2: High Quality Token Expander}
\paragraph{How to Generate High-Quality Tokens?}
Although large-scale pruning of visual tokens in Stage 1 helps reduce hallucination by removing semantically irrelevant inputs, it also introduces a potential risk: some tokens that are globally low in attention may still carry locally important visual details and could be mistakenly discarded. These tokens may contain fine-grained information such as edge structures, small objects, or background evidence that contribute to overall scene understanding. Relying solely on global attention for hard filtering can thus lead to semantic degradation.

To address this issue, we introduce a similarity-guided visual token enhancement mechanism in Stage 2. Our method is inspired by classical feature engineering pipelines, particularly those based on the expand-and-compress paradigm such as OpenFE~\citep{zhang2023openfe}. In these frameworks, new features are first expanded from base attributes through composition and then selectively compressed via performance-driven ranking or pruning (in Stage 3). This process allows models to explore a richer set of representations while maintaining computational tractability. Motivated by this, we propose to treat tokens as features, and design a multi-stage editing framework that performs explicit token-level expansion and compression within large vision-language models. In this view, visual tokens serve as raw features extracted from images; we then enhance and refine their semantic structure by applying multiple rounds of selective retention and similarity-guided fusion. This process, which we term token engineering, reshapes the token space into a more structured and expressive representation.

By combining expansion with targeted selection, our method not only removes redundancy but also generates high-quality, semantically enriched tokens that better support downstream inference and alignment.
This module leverages local semantic similarities among the original visual tokens to generate fused representations that recover potentially useful information lost during aggressive pruning. These enriched tokens serve as structural complements to reinforce visual grounding in the decoding process.

\paragraph{Why Use Original Tokens for Fusion?}
A natural question arises: if Stage 1 causes information loss, why does Stage 2 fuse tokens from the \textit{original} embedding layer rather than directly recovering the pruned tokens? We argue that the pruned tokens are discarded precisely because they have low cross-modal relevance—re-introducing them directly would re-inject noise. Instead, Stage 2 takes a different approach: it generates \textit{new} high-quality tokens by fusing semantically similar original tokens. These fused tokens serve as a ``semantic buffer'' that may capture fine-grained visual evidence missed by the attention-based pruning. Stage 3 then selects from this enriched pool based on textual alignment, ensuring that only genuinely useful information is retained. Our ablation study in \cref{tab:ablation_main} validates this design: Stage 2's similarity-guided fusion achieves comparable or better recovery than naively appending original tokens, while being more efficient.

\paragraph{Token Expansion}
Let $X \in \mathbb{R}^{B \times k \times d}$ denote the original visual tokens set in the embedding layer, where $B$ is the batch size, $k = L_v$ is the number of visual tokens, and $d$ is the dimension of features. We first apply $\ell_2$ normalization to each token along the feature dimension and compute the cosine similarity matrix between all token pairs:
\[
S = \left( \frac{X}{\|X\|_2} \right) \cdot \left( \frac{X}{\|X\|_2} \right)^\top \in \mathbb{R}^{B \times k \times k},
\]

To avoid self-similarity, we mask the diagonal entries of $S$ with $-\infty$. Then, for each token, we retrieve its top-$t$ most similar neighbors. Let $x_i$ be the $i$-th token, and $N_i = \{x_{j_1}, \ldots, x_{j_t}\}$ be its top-$t$ neighbors. We compute the fusion result by:
\[
\hat{x}_i^{(j)} = \alpha_{ij} \cdot x_i + (1 - \alpha_{ij}) \cdot x_j, \quad
\text{where } \alpha_{ij} = \frac{s_{ij}}{s_{ij} + 1}
\]

where $s_{ij}$ is the similarity between token $x_i$ and its neighbor $x_j$. All $\hat{x}_i^{(j)}$ are concatenated and reshaped into enhanced $k \cdot t$ tokens. These enhanced tokens form an enriched visual representation and serve as input to the next stage, where global attention is used to select the most semantically aligned tokens. In this way, the enriched representation acts as an intermediate buffer that supports a more robust cross-modal grounding in Stage 3.

\begin{table*}[t]
\centering
\caption{Experimental results of various decoding strategies on MME dataset across four models: LLaVA-1.5, LLaVA-NEXT, INF-MLLM and mPLUG-Owl2. The best values are highlighted in \textbf{bold}.}
\label{tab:mme}
\resizebox{\textwidth}{!}{%
\begin{tabular}{l|ccc|ccc|ccc|ccc}
\toprule
\multirow{2}{*}{\textbf{Decoding}}
& \multicolumn{3}{c|}{\textbf{LLaVA-1.5}}
& \multicolumn{3}{c|}{\textbf{LLaVA-NEXT}}
& \multicolumn{3}{c|}{\textbf{INF-MLLM}}
& \multicolumn{3}{c}{\textbf{mPLUG-Owl2}}
\\
\cmidrule(lr){2-4} \cmidrule(lr){5-7}\cmidrule(lr){8-10}\cmidrule(lr){11-13}
& Perc. & Cog. & Total & Perc. & Cog. & Total & Perc. & Cog. & Total & Perc. & Cog. & Total \\
\midrule
Regular & 1491.56 & 294.29 & 1785.85 & 1519.30 & 330.00 & 1849.30 & 1491.96 & 266.07 & 1758.03 & 1459.54 & 345.71 & 1805.25 \\
VCD  & 1484.96 & 287.50 & 1772.46 & 1418.27 & \textbf{351.07} & 1769.34 & 1444.36 & 270.71 & 1715.07 & 1311.52 & 329.29 & 1640.81 \\
DoLa & 1495.02 & \textbf{318.21} & 1813.23 & 1515.41 & 262.14 & 1777.55 & 1491.15 & 265.00 & 1756.15 & 1462.33 & 265.00 & 1722.33 \\
DCLA & \textbf{1520.14} & 280.00 & 1800.14 & 1525.73 & 330.00 & 1855.73 & 1509.05 & \textbf{273.21} & \textbf{1782.26} & 1463.40 & 334.29 & 1797.69 \\
SPIN & 1491.91 & 295.71 & 1787.62 & 1506.37 & 326.07 & 1832.44 & 1493.21 & 268.33 & 1761.54 & 1465.38 & 332.56 & 1797.94 \\
\textbf{SeeMe} & 1519.49 & 310.00 & \textbf{1829.49} & \textbf{1527.34} & 346.43 & \textbf{1873.77} & \textbf{1511.17} & 269.29 & 1780.46 & \textbf{1472.04} & \textbf{345.71} & \textbf{1817.75} \\
\bottomrule
\end{tabular}%
}
\vspace{-2mm}
\end{table*}

\subsection{Stage 3: Cross-Modal Token Refiner}
\subsubsection{Stage 3 is Essential for Hallucination Suppression}
\begin{figure}[t]
\centering
\includegraphics[width=0.9\linewidth]{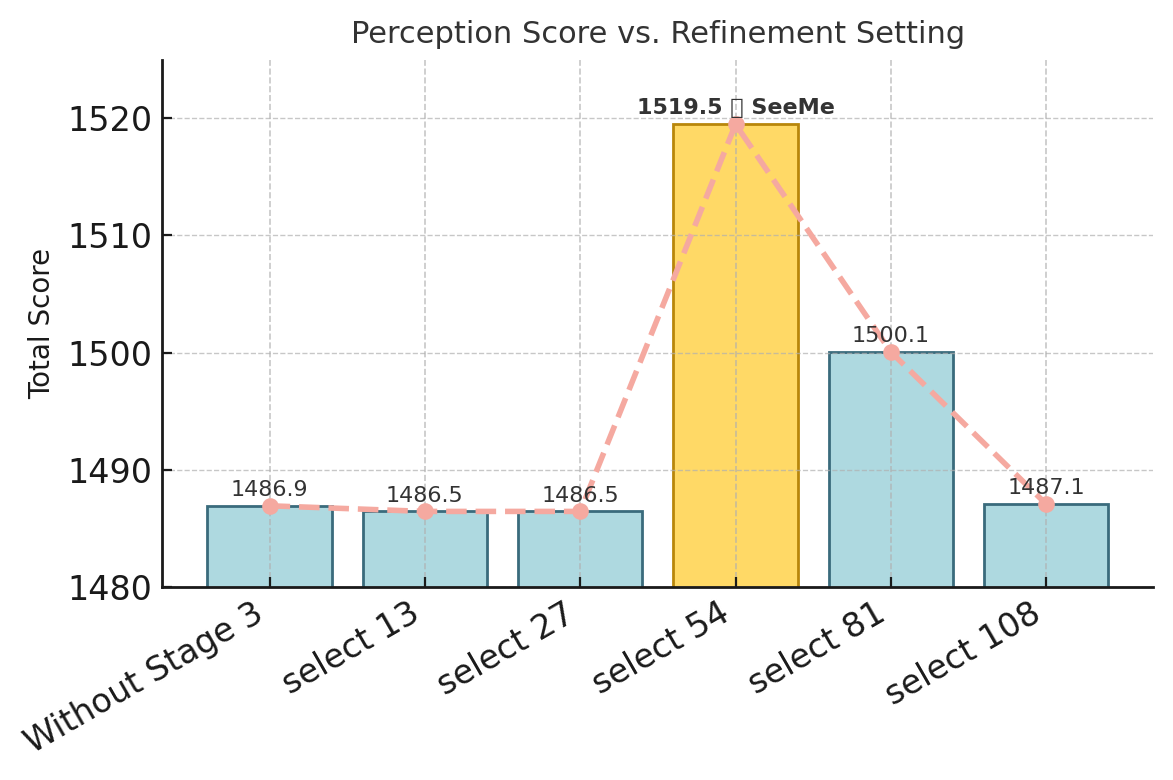}
\caption{This figure shows that both omitting the final refinement stage (Stage 3) and selecting suboptimal token counts (too few or too many) lead to lower overall scores. Selecting 54 tokens in Stage 3 yields the highest performance, validating its necessity.}
\label{fig_stage3}
\end{figure}

\cref{fig_stage3} illustrates the impact of Stage 3 on perception performance across different refinement settings. It is evident that simply retaining all tokens (Without Stage 3) does not effectively suppress hallucination. However, overly aggressive pruning leads to information loss, while insufficient pruning retains excessive noise and redundancy. Only the reasonable setting—corresponding to our proposed SeeMe method—achieves a favorable trade-off, significantly enhancing performance. These results highlight the necessity of stage 3 as a crucial step for balancing semantic retention and hallucination suppression.

\subsubsection{Cross-Modal Token Refiner}
While the fused tokens generated in Stage 2 help recover fine-grained local information lost during pruning, not all of them are equally relevant to the textual context. To further enhance semantic alignment, we introduce a cross-modal refinement mechanism that selects the most text-aligned fused tokens using global self-attention.

We feed the entire sequence into the decoder layer $K$  and extract its self-attention map $A \in \mathbb{R}^{B \times H \times L \times L}$, where $B$ is the batch size, $H$ is the number of attention heads, and $L$ is the total length of the sequence. For each fused token $i$, we compute its average attention received from all textual tokens:
\[
\text{CrossScore}(i) = \frac{1}{L_q + L_o} \sum_{j=1}^{L_q + L_o} A[:, :, j, i] \in \mathbb{R}^{B \times H}.
\]

We average across all heads and the batch dimension to obtain a single scalar score per fused token, and select the top $n$ tokens with the highest scores. These selected fused tokens are then concatenated with the retained visual tokens from Stage 1. This global selection step refines the visual representation by retaining only those fused tokens that are highly aligned with the language stream. It enhances grounding accuracy while suppressing residual hallucinations introduced by low-confidence visual content.

\begin{table*}[t]
\centering
\caption{Experimental results of various decoding strategies on POPE dataset across four models: LLaVA-1.5, LLaVA-NEXT, INF-MLLM and mPLUG-Owl2. We used the average accuracy and F1 score of the Random, Popular and Adversarial splits. The best values are highlighted in \textbf{bold}.}
\label{tab:pope}
\small
\setlength{\tabcolsep}{10pt}
\setlength{\textfloatsep}{2pt}
\begin{tabular}{l|l|cc|cc|cc}
\toprule
\multirow{2}{*}{\textbf{Model}}
& \multicolumn{1}{c|}{}
& \multicolumn{2}{c|}{\textbf{\textit{MSCOCO}}}
& \multicolumn{2}{c|}{\textbf{\textit{A-OKVQA}}}
& \multicolumn{2}{c}{\textbf{\textit{GQA}}} \\
\cmidrule(lr){2-8}
& \textbf{Decoding} & \textbf{Accuracy} & \textbf{F1 Score}& \textbf{Accuracy} & \textbf{F1 Score} & \textbf{Accuracy} & \textbf{F1 Score} \\
\midrule
\multirow{6}{*}{LLaVA-1.5}
& Regular & 85.19 & 86.10& 78.84 & 82.51 & 76.57 & 80.98 \\
& VCD & 84.66 & 85.35 & 77.86 & 81.41 & 75.94 & 80.36 \\
& DoLa & 85.08 & 85.99 & 78.70 & 82.28 & 76.56 & 80.84 \\
& DCLA & 85.82 & 86.45 & \textbf{79.76} & \textbf{83.12} & 77.41 & 81.51 \\
& SPIN & 85.34 & 86.12 & 79.41 & 82.86 & 77.22 & 81.40 \\
& \textbf{SeeMe} & \textbf{86.07} & \textbf{86.73} & 79.66 & 82.99 & \textbf{77.50} & \textbf{81.52} \\
\midrule
\multirow{6}{*}{LLaVA-NEXT}
& Regular & 87.62 & 86.42 & 87.41 & 87.46 & 86.58 & 86.41 \\
& VCD & 79.65 & 74.82 & 79.24 & 75.80 & 78.85 & 75.27 \\
& DoLa & 84.91 & 82.55 & 86.46 & 85.63 & 84.64 & 83.20 \\
& DCLA & \textbf{87.71} & \textbf{86.49} & 87.55 & 87.58 & 86.61 & 86.41 \\
& SPIN & 86.89 & 85.35 & 87.11 & 87.06 & 85.99 & 85.67 \\
& \textbf{SeeMe} & 87.62 & 86.44 &\textbf{87.61} & \textbf{87.77} & \textbf{86.62} & \textbf{86.47} \\
\midrule
\multirow{6}{*}{INF-MLLM}
& Regular & 88.29 & 88.36 & 83.66 & 84.37 & 82.77 & 84.73 \\
& VCD & 85.56 & 85.73 & 81.70 & 83.70 & 79.79 & 82.05 \\
& DoLa & 88.28 & 88.36 & 84.12 & 85.89 & 82.88 & 84.81 \\
& DCLA & 88.43 & 88.46 & 84.14 & 85.91 & 83.04 & 84.94 \\
& SPIN  & 88.31 & 88.41  & 83.46 & 84.33  & 83.47 & 84.98\\
& \textbf{SeeMe} &\textbf{88.47} & \textbf{88.56} & \textbf{84.70} & \textbf{86.01} &\textbf{ 83.67} & \textbf{85.01} \\
\midrule
\multirow{6}{*}{mPLUG-Owl2}
& Regular & 86.39 & 85.91 & 83.03 & 83.69 & 81.18 & 81.61 \\
& VCD & 79.03 & 78.56 & 79.03 & 78.56 & 79.03 & 78.00 \\
& DoLa & 85.98 & \textbf{86.34} & 83.42 & 83.42 & \textbf{83.49} & 80.38 \\
& DCLA & 86.51 & 86.19 & 83.20 & 84.31 & 81.61 & 81.94 \\
& SPIN & 86.41 & 86.03  & 83.43 & 84.25& 82.11 & 81.78\\
& \textbf{SeeMe} & \textbf{86.77} & 86.32 & \textbf{83.76} & \textbf{84.42} & 82.33 & \textbf{82.27}\\
\bottomrule
\end{tabular}
\vspace{-3mm}
\end{table*}

\section{Experiment}
\subsection{Setup}
We apply SeeMe with model-specific configurations across different LVLMs. For all models, attention-based cross-modal pruning is applied at a single decoder layer (without additional training), with a certain ratio of visual tokens retained. In the merging stage, each token is fused with its top $k$ most similar tokens using cosine similarity. In the final selection stage, a fixed number of visual tokens is retained based on alignment with the linguistic context. The specific parameter design is reported in \cref{sec:hyperparameter_settings} and \cref{tab:hyperparameter_settings}.

For other decoding methods, we uniformly set the temperature to 0 for the fairness of the experiment. The specific parameters are presented in \cref{sec:hyperparameter_settings}.

\paragraph{Datasets} To thoroughly evaluate the effectiveness of our SeeMe method in addressing hallucination issues in Large Vision-Language Models(LVLMs), we employed the MME benchmark \citep{Fu2023MME}, which includes 14 tasks categorized into perception and cognition. Additionally, we focused on assessing SeeMe's performance in mitigating object hallucinations using the POPE benchmark (Polling-based Object Probing Evaluation) \citep{li2023evaluating}, which utilizes SEEM-annotated datasets such as MSCOCO \citep{lin2014microsoft}, A-OKVQA \citep{schwenk2022okvqa} and GQA \citep{hudson2019gqa}. Furthermore, we used AMBER \citep{wang2023amber} to evaluate the effectiveness of the SeeMe method in suppressing model hallucinations in multimodal discrimination tasks. AMBER is a multi-dimensional benchmark dataset designed for LVLMs to evaluate the impact of different types of hallucinations (such as presence hallucinations, attribute hallucinations, and relational hallucinations) on model performance.

\paragraph{Models and Baselines} We conducted experiments on four recent LVLMs with 7B parameters: LLaVA-1.5 \citep{liu2023llava}, LLaVA-NEXT \citep{liu2024llavanext}, INF-MLLM \citep{zhou2023infmllm}, and mPLUG-Owl2 \citep{ye2024mplug}. These models are commonly used in LVLM benchmarks and vary significantly in their vision-language fusion approaches and pre-training strategies, offering a robust framework for testing the applicability of our method.

For baseline comparisons, we evaluated SeeMe against several prominent decoding techniques, including standard decoding, contrastive decoding methods such as VCD \citep{leng2024mitigating} and DoLa \citep{chuang2023dola}, an inter-layer mechanism: DCLA \citep{tang2026mitigatinghallucinationsinterlayerconsistency}, and an image-guided inference approach: SPIN \citep{sarkar2025mitigating}. These methods were selected due to their representation of different conceptual approaches in the current landscape of decoding optimization: VCD and DoLa focus on enhancing decoding strategies, DCLA prioritizes maintaining semantic consistency across layers, and SPIN leverages visual input to guide inference, thus addressing a range of mainstream optimization directions. To ensure fairness and reproducibility, all decoding strategies were evaluated under the same conditions, with the decoding temperature consistently set to zero across all experiments.

\subsection{Results}
\paragraph{Result on MME} We evaluated the performance of SeeMe and several decoding strategies across four models: LLaVA-1.5, LLaVA-NEXT, INF-MLLM, and mPLUG-Owl2, with the experimental results summarized in \cref{tab:mme}. The table presents the Perception, Cognition and Total scores for each model under different decoding strategies. In the LLaVA-1.5, LLaVA-NEXT and mPLUG-Owl2 models, SeeMe outperformed regular decoding, VCD, DoLa, DCLA, and SPIN methods, achieving total scores of 1829.49, 1873.77, and 1817.75, respectively. In the INF-MLLM model, SeeMe also demonstrated strong performance, attaining a total score of 1780.46, significantly surpassing most other decoding strategies. These results clearly highlight the advantages of SeeMe in improving decoding performance and mitigating hallucinations.

\paragraph{Result on POPE} To evaluate the effectiveness of our proposed method in mitigating object-level hallucinations, we conducted experiments on the SEEM-annotated versions of the \textit{MSCOCO}, \textit{A-OKVQA}, and \textit{GQA} datasets, which are part of the POPE benchmark. The POPE benchmark is widely used for assessing hallucinations in visual question answering (VQA) tasks, focusing on the accuracy and reliability of generated object mentions in image captions. The benchmark includes three distinct splits: Random Split (any object from the dataset), Popular Split (the most frequent objects in the dataset), and Adversarial Split (objects that are closely related but misleading). These splits represent different levels of challenge, from general object recognition to dealing with tricky or misleading objects that may cause hallucinations. In this study, we compare the performance of SeeMe with several representative decoding strategies, including regular decoding, VCD, DoLa, DCLA and SPIN. For each of the three splits, we calculate the average score across all categories and use it as the final evaluation score to assess the overall effectiveness of the decoding strategies in reducing hallucinations.

As shown in \cref{tab:pope}, SeeMe achieved high accuracy and F1 scores on the \textit{MSCOCO}, \textit{A-OKVQA}, and \textit{GQA} datasets across the LLaVA-1.5, LLaVA-NEXT, INF-MLLM, and mPLUG-Owl2, surpassing other decoding strategies and demonstrating substantial improvements. These results highlight SeeMe's ability to consistently enhance model performance, particularly in terms of inference accuracy and effectively mitigating hallucinations in LVLMs.

\paragraph{Results on AMBER} To thoroughly evaluate the effectiveness of our SeeMe method in addressing hallucination issues in Large Vision-Language Models (LVLMs), we employed the AMBER benchmark, which offers a comprehensive framework for hallucination evaluation. AMBER provides a multi-dimensional evaluation that focuses on three primary types of hallucinations: existence, attribute, and relation hallucinations, making it a valuable tool for assessing the impact of hallucinations on model performance. We use discriminative tasks in AMBER to examine SeeMe's ability to mitigate hallucinations by improving the model's judgment of object existence, attributes and relationships, ensuring the alignment of model responses with actual visual content. The experimental data in \cref{tab:amber} indicates that, compared to other decoding strategies, SeeMe achieved the best results across LLaVA-1.5, LLaVA-NEXT, INF-MLLM and mPLUG-Owl2. Overall, SeeMe exhibits superior anti-hallucination capability and discrimination accuracy, thereby validating its effectiveness in mitigating hallucinations in multimodal discrimination tasks.

\begin{table}[t]
\vspace{-0.8mm}
\centering
\small
\caption{Experimental results on discriminative tasks from AMBER dataset across four models: LLaVA-1.5, LLaVA-NEXT, INF-MLLM and mPLUG-Owl2. We used the overall accuracy and F1 score of discriminative tasks. The best values are highlighted in \textbf{bold}.}
\label{tab:amber}
\setlength{\tabcolsep}{10pt}
\begin{tabular}{l|l|cc}
\toprule
\multirow{1}{*}{\textbf{Model}}
& \textbf{Decoding} & \textbf{Accuracy} & \textbf{F1 Score} \\
\midrule
\multirow{6}{*}{LLaVA-1.5}
& Regular & 71.5 & 74.1 \\
& VCD & 72.0 & 74.9 \\
& DoLa & 71.5 & 74.2 \\
& DCLA & 72.6 & 75.7 \\
& SPIN & 72.3 & 75.3\\
& \textbf{SeeMe}  & \textbf{72.7} & \textbf{75.9} \\
\cmidrule(lr){1-4}
\multirow{6}{*}{LLaVA-NEXT}
& Regular & 83.6 & 87.7 \\
& VCD & 82.6 & 87.1 \\
& DoLa & 83.3 & 87.7 \\
& DCLA & 83.5 & 87.7 \\
& SPIN & 83.1 & 87.5\\
& \textbf{SeeMe}  & \textbf{83.7} & \textbf{87.8} \\
\cmidrule(lr){1-4}
\multirow{6}{*}{INF-MLLM}
& Regular & 71.4 & 74.3 \\
& VCD & 71.5 & 74.4 \\
& DoLa & 72.2 & 74.8 \\
& DCLA & 72.7 & 75.1 \\
& SPIN & 70.9 & 74.1\\
& \textbf{SeeMe}  & \textbf{73.4} & \textbf{75.1} \\
\cmidrule(lr){1-4}
\multirow{6}{*}{mPLUG-Owl2}
& Regular & 76.3 & 78.8 \\
& VCD & 75.9 & 78.4 \\
& DoLa & 76.1 & 79.1 \\
& DCLA & 76.5 & 79.3 \\
& SPIN & 76.4 & 78.8\\
& \textbf{SeeMe}  &\textbf{76.8 }& \textbf{79.7}\\
\bottomrule
\end{tabular}
\vspace{-3mm}
\end{table}
\subsection{Ablation Study}
We conducted a comprehensive ablation study to analyze the contributions of different components in SeeMe's three-stage pipeline. As shown in \cref{tab:ablation_main}, we examine the effect of each stage under different retain ratios.

\begin{table}[h]
\centering
\small
\setlength{\tabcolsep}{5pt}
\caption{Ablation study on MME at decoder layer 10. Pure pruning causes information loss; appending original tokens recovers it. Stage 2 achieves similar recovery, and Stage 3 further refines alignment.}
\label{tab:ablation_main}
\begin{tabular}{lcccc}
\toprule
\textbf{Method} & \textbf{r=0.5} & \textbf{r=0.25} & \textbf{r=0.1} & \textbf{r=0.05} \\
\midrule
A: S1 only                & 1496.3 & 1476.3 & 1411.1 & 1325.4 \\
B: A + Orig. tokens       & 1496.4 & 1483.4 & 1488.3 & 1473.7 \\
C: S1 + S2                & 1497.7 & 1486.7 & 1490.5 & 1473.6 \\
D: SeeMe (Full)           & \textbf{1499.8} & \textbf{1489.6} & \textbf{1492.4} & \textbf{1478.1} \\
\midrule
\multicolumn{5}{l}{\textit{Key comparisons:}} \\
$\Delta$(B$-$A): Orig. tokens  & +0.1 & +7.1 & +77.3 & +148.3 \\
$\Delta$(C$-$A): S2 recovery   & +1.4 & +10.4 & +79.4 & +148.2 \\
$\Delta$(C$-$B): S2 vs append  & +1.3 & +3.3 & +2.2 & $-$0.1 \\
$\Delta$(D$-$C): S3 refine     & +2.1 & +2.1 & +2.0 & +4.5 \\
\bottomrule
\end{tabular}
\vspace{-3mm}
\end{table}

The results reveal several key findings: (1) Pure pruning (Stage 1 only) causes significant information loss, especially at low retain ratios ($\downarrow$166 at r=0.05); (2) Appending original embedding tokens recovers most of the lost performance, proving that pruned tokens contain valuable visual semantics; (3) Stage 2's similarity-guided fusion achieves comparable or better recovery than naive appending, while being more efficient; (4) Stage 3 consistently provides additional refinement (+2$\sim$4 points) through cross-modal alignment. A more detailed layer/ratio analysis is provided in \cref{sec:extended_ablation}.

\subsection{Efficiency Analysis}
We analyze the computational overhead of different decoding strategies on LLaVA-1.5-7B. As shown in \cref{tab:efficiency}, SeeMe achieves the lowest prefill latency (45.56 ms/token), while maintaining comparable decode latency (22.73 ms/token). More importantly, SeeMe attains the highest throughput (9.23 images/s) among all methods, while not incurring significant memory overhead compared to regular decoding. These results demonstrate that SeeMe's three-stage pipeline introduces negligible computational cost while delivering superior hallucination mitigation, making it highly practical for real-world deployment.

\begin{table}[t]
\centering
\small
\setlength{\tabcolsep}{3pt}
\caption{Efficiency comparison of different decoding strategies on LLaVA-1.5-7B. $\downarrow$ indicates lower is better; $\uparrow$ indicates higher is better. The best values are highlighted in \textbf{bold}.}
\label{tab:efficiency}
\begin{tabular}{lcccc}
\toprule
& \scriptsize\textbf{Prefill Latency}$\downarrow$ & \scriptsize\textbf{Decode Latency}$\downarrow$ & \scriptsize\textbf{GPU Usage}$\downarrow$ & \scriptsize\textbf{Throughput}$\uparrow$ \\
& \scriptsize\textbf{(ms/token)} & \scriptsize\textbf{(ms/token)} & \scriptsize\textbf{(GB)} & \scriptsize\textbf{(Images/s)} \\
\midrule
Regular & 50.46 & \textbf{21.65} & 14.72 & 9.14 \\
VCD & 174.18 & 174.65 & \textbf{14.04} & 2.85 \\
DoLa  & 135.11 & 130.52 & 14.92 & 1.39 \\
DCLA & 67.46 & 30.08 & 18.77 & 7.45 \\
SPIN & 74.88 & 22.19 & 15.27 & 7.80 \\
\textbf{SeeMe}  & \textbf{45.56} & 22.73 & 15.88 & \textbf{9.23} \\
\bottomrule
\end{tabular}
\vspace{-3mm}
\end{table}

\subsection{Hyperparameter Settings}
\label{sec:hyperparameter_settings}
For the MME and POPE benchmarks, SeeMe is activated at the 14th decoder layer of LLaVA-1.5 and mPLUG-Owl2, at the 16th layer of LLaVA-NEXT, and at the 13th layer of INF-MLLM, each with a pruning retain ratio of 0.05. During merging, each token is fused with its most similar neighbors: top-2 by cosine similarity for LLaVA-1.5, INF-MLLM, and mPLUG-Owl2, and top-1 for LLaVA-NEXT. The final number of selected tokens ($n$) is fixed at 54 for LLaVA-1.5 and mPLUG-Owl2, and 27 for INF-MLLM and LLaVA-NEXT. When evaluating on AMBER, we adjust the retain ratio to 0.01.

For all other methods, we use the official open-source parameters, while AMBER retains the same parameters as the other datasets. As a special case, we explicitly set the temperature of VCD to 0 to ensure experimental fairness. SPIN uses its POPE parameters when evaluated on AMBER.

\begin{table}[t]
  \centering
  \footnotesize
  \setlength{\tabcolsep}{2pt}
  \caption{Hyperparameter settings of SeeMe for each model.}
  \label{tab:hyperparameter_settings}
  \begin{tabular}{lcccc}
    \toprule
          & LLaVA-1.5 & LLaVA-NEXT & INF-MLLM & mPLUG-Owl2 \\
    \midrule
    $K$ & 14 & 16 & 13 & 14  \\
    ratio & 0.05 & 0.05 & 0.05 & 0.05  \\
    top-$k$ & 2 & 1 & 2 & 2  \\
    select-$n$ & 54 & 27 & 27 & 54  \\
    \bottomrule
  \end{tabular}
  \vspace{-3mm}
\end{table}

\paragraph{Hyperparameter Selection Guidelines.}
To facilitate the application of SeeMe to new models, we provide practical guidelines for hyperparameter selection. We recommend selecting $K$ around 40--50\% of the total decoder layers. For a 32-layer model, $K \in [13, 16]$ typically works well. The key insight is that cross-modal attention should have matured, meaning that visual-textual alignment has been established, but should not yet be diluted by overly dispersed attention. Pruning too early loses relevant tokens, while pruning too late misses the opportunity to filter noise before it propagates.

A retain ratio of 0.05 works robustly across all tested models. Lower ratios, such as 0.01, may improve hallucination suppression but risk information loss, which Stage 2 and Stage 3 must compensate for. Higher ratios, such as 0.25, are safer but provide less noise reduction. For the merging and final selection stages, top-$k \in \{1, 2\}$ and select-$n \approx 0.1 \times |\mathbf{Z}_v|$ provide a good balance. For example, selecting 54 tokens works well when the input contains 576 visual tokens. Larger $k$ generates more fused tokens but increases redundancy, while smaller $n$ is more aggressive and may discard useful fused tokens.

For a new model, we suggest starting with $K = \lfloor 0.45 \times \text{num\_layers} \rfloor$, ratio $=0.05$, top-$k=2$, and select-$n=54$. A small validation set, such as 100 samples from MME, can then be used to adjust $K$ by $\pm 2$ layers and tune the ratio or select-$n$ according to the trade-off between hallucination suppression and information retention.

\subsection{Extended Ablation Study}
\label{sec:extended_ablation}
As shown in \cref{fig_motivation}, the layer/retain-ratio sweep measures how aggressively to apply cross-modal pruning on the MME perception task. We sweep the retain ratio of visual tokens (50\%, 25\%, 10\%, 5\%) at decoder layers 10--17 in LLaVA-1.5-7B, using the regular decoding score of 1491.56 as the baseline. The results show that pruning in intermediate layers, especially layers 12--14, consistently exceeds the baseline, whereas pruning too early or too late is less stable.

\section{Conclusion}
In this paper, we identify that redundant and noisy visual tokens can mislead LVLMs, leading to output that is inconsistent with the actual visual information. To address this issue, we proposed \textbf{SeeMe}, the first method to integrate the concept of feature engineering from traditional machine learning into LVLMs, offering an innovative solution to the hallucination problem. Through a three-stage visual tokens reconstruction process, SeeMe effectively reduces redundancy and noise in visual tokens without requiring additional training, minimizing the inconsistency between the generated content and the actual visual input, and enhancing the overall reliability and effectiveness of the model. Extensive experiments across multiple benchmark datasets demonstrate the effectiveness of SeeMe in various multimodal tasks, particularly in visual question answering, where it significantly improves the accuracy and stability of the generated results.

\section*{Impact Statement}

This paper presents work whose goal is to advance the field of Machine Learning, specifically in improving the reliability of Large Vision-Language Models by mitigating hallucinations. Our method, SeeMe, is a training-free approach that enhances the accuracy of visual understanding without requiring additional data or computational resources for retraining. The potential societal benefits include more reliable AI systems for applications such as medical image analysis, autonomous driving, and assistive technologies. We do not foresee any specific negative societal consequences that must be highlighted here, though we encourage responsible deployment and continued research into AI safety and reliability.

\bibliography{seeme}
\bibliographystyle{icml2026}

\end{document}